\documentclass[journal,twoside,web]{ieeecolor}
\usepackage{tmi}
\usepackage{cite}
\usepackage{amsmath,amssymb,amsfonts}
\usepackage{algorithmic}
\usepackage{graphicx}
\usepackage{textcomp}
\usepackage{multirow}
\usepackage{url}
\def\BibTeX{{\rm B\kern-.05em{\sc i\kern-.025em b}\kern-.08em
    T\kern-.1667em\lower.7ex\hbox{E}\kern-.125emX}}
\begin{document}
\title{XBound-Former: Toward Cross-scale Boundary Modeling in Transformers}
\author{Jiacheng Wang, Fei Chen, Yuxi Ma, Liansheng Wang, Zhaodong Fei, Jianwei Shuai, Xiangdong Tang, Qichao Zhou, Jing Qin
\thanks{Jiacheng Wang is with the Department of Computer Science at School of Informatics, Xiamen University, Xiamen 361005, China. He finished this work during the intern of Manteia Technologies Co.,Ltd. (e-mail:jiachengw@stu.xmu.edu.cn)}
\thanks{Fei Chen, Liansheng Wang are with the Department of Computer Science at School of Informatics, Xiamen University, Xiamen 361005, China. (e-mail:feichen@stu.xmu.edu.cn, lswang@xmu.edu.cn)}
\thanks{Yuxi Ma is with the Xiamen University, Xiamen 361005, China. (e-mail:mayuxi1@stu.xmu.edu.cn)}
\thanks{Zhaodong Fei is with the Department of Radiotherapy, Fujian Cancer Hospital, Fuzhou 350000, China (e-mail:feizhaodong@yeah.net)}
\thanks{Jianwei Shuai is with the Oujiang Laboratory (Zhejiang Lab for Regenerative Medicine, Vision and Brain Health), and Wenzhou Institute, University of Chinese Academy of Sciences, Wenzhou, Zhejiang 325001, China. (jianweishuai@xmu.edu.cn)}
\thanks{Xiangdong Tang is with the Sleep Medicine Center, Mental Health Center, Department of Respiratory and Critical Care Medicine, State Key Laboratory of Biotherapy, West China Hospital, Sichuan University, Chengdu 610041, China. (2372564613@qq.com)}
\thanks{Qichao Zhou is with Manteia Technologies Co.,Ltd. (e-mail:zhouqc@manteiatech.com)}
\thanks{Jing Qin is with Center for Smart Health, School of Nursing, The Hong Kong Polytechnic University. (e-mail:harry.qin@polyu.edu.hk)}
\thanks{Liansheng Wang and Qichao Zhou are the corresponding authors.}
}

\maketitle

\begin{abstract}
Skin lesion segmentation from dermoscopy images is of great significance in the quantitative analysis of skin cancers, which is yet challenging even for dermatologists due to the inherent issues, i.e., considerable size, shape and color variation, and ambiguous boundaries.
Recent vision transformers have shown promising performance in handling the variation through global context modeling. Still, they have not thoroughly solved the problem of ambiguous boundaries as they ignore the complementary usage of the boundary knowledge and global contexts.
In this paper, we propose a novel cross-scale boundary-aware transformer, \textbf{XBound-Former}, to simultaneously address the variation and boundary problems of skin lesion segmentation.
\textbf{XBound-Former} is a purely attention-based network and catches boundary knowledge via three specially designed learners.
First, we propose an implicit boundary learner (im-Bound) to constrain the network attention on the points with noticeable boundary variation, enhancing the local context modeling while maintaining the global context.
Second, we propose an explicit boundary learner (ex-Bound) to extract the boundary knowledge and convert it into embeddings explicitly.
We learn the knowledge at different scales, offering multi-scale perspectives to exploit the boundary representations.
Third, based on the learned multi-scale boundary embeddings, we propose a cross-scale boundary learner (X-Bound) to simultaneously address the problem of ambiguous and multi-scale boundaries by using learned boundary embedding from one scale to guide the boundary-aware attention on the other scales.
We evaluate the model on two skin lesion datasets, ISIC-2016\&PH$^2$ and ISIC-2018, where our model consistently outperforms other convolution- and transformer-based models, especially on the boundary-wise metrics.
We extensively verify the generalization ability of polyp lesion segmentation that has similar characteristics, and our model can also yield significant improvement compared to the latest models.
All resources could be found in \url{https://github.com/jcwang123/xboundformer}.
\end{abstract}


\section{Introduction}
\begin{figure}[t]
    \centering
    \includegraphics[width=\linewidth]{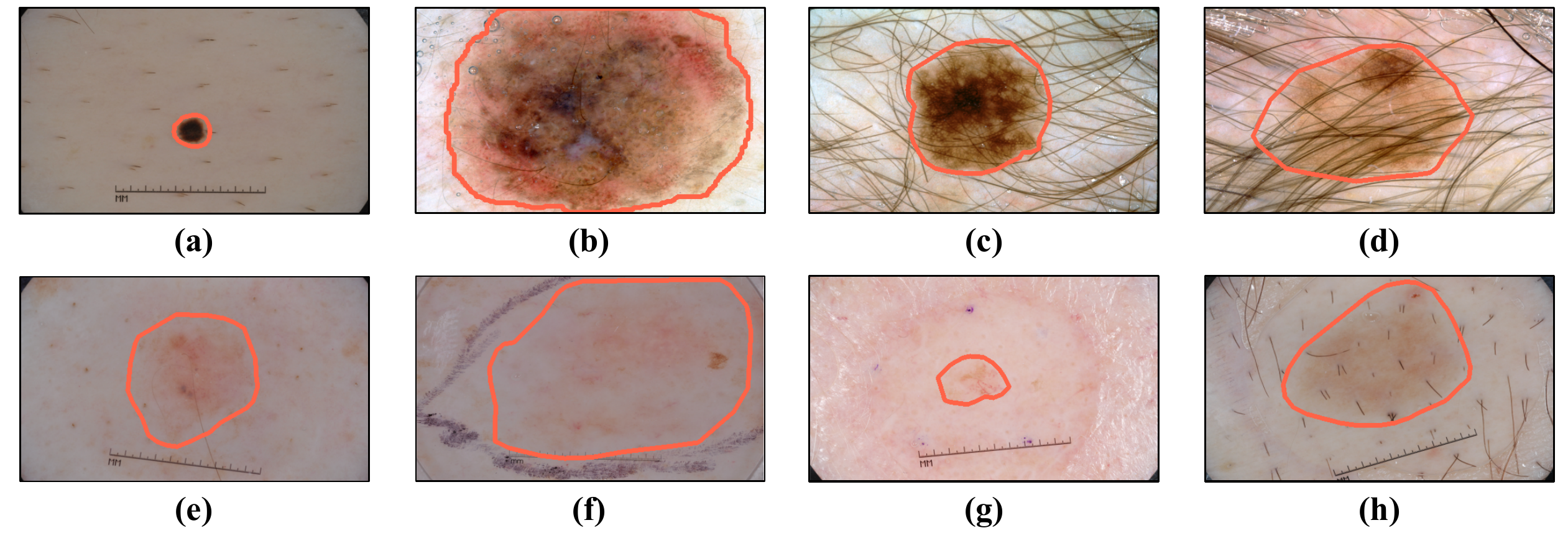}
    \caption{The challenges of automatic skin lesion segmentation from  dermoscopy images: (a)-(b) large skin lesion variations in size, shape, and color, (c)-(d) partial occlusion by hair, and (e)-(h) ambiguous boundaries.
    }
    \label{dataset}
    \vspace{-4mm}
\end{figure}

Melanoma is one of the most rapidly increasing cancers over the world, consistently leading to about 100,000 new cases and 7000 deaths per year~\cite{mathur2020cancer,siegel2021cancer}.
Segmenting skin lesions from dermoscopy images is critical in the diagnosis and treatment planning, which is usually tedious, time-consuming, and error-prone for human beings.
In this regard, automated segmentation methods are highly demanded in clinical practice to improve clinical workflow in terms of accuracy and efficiency.
It remains a very challenging task because
(1) skin lesions have large size, shape and color variance (see Fig.~\ref{dataset} (a-b)),
(2) the hair will partially cover the lesions destroying the local context (see Fig.~\ref{dataset} (c-d)),
(3) sometimes, the contrast between lesions to normal skin is relatively low, resulting in ambiguous boundaries (see Fig.~\ref{dataset} (e-h)).

Many efforts have been dedicated to overcoming these challenges.
Hand-crafted features are adopted in the early years, which are usually not stable and robust, leading to poor segmentation performance when facing lesions with large variations~\cite{5342426}.
To solve this problem, deep learning models based on convolutional neural networks (CNN) have been proposed and achieved remarkable performance gains~\cite{yuan2017automatic,7792699}.
However, due to the lack of global context modeling, these models are still insufficient to counteract the large variation of skin lesion segmentation.
Concerning this, researchers propose various approaches to enlarge the receptive fields inspired by the advancement of dilated convolution~\cite{yu2017dilated,yu2019multi}.
Lee~\textit{et al.}~\cite{9157193} extensively incorporate the dilated attention module with boundary prior so that the network predicts boundary key-points maps to guide the attention module.

Nevertheless, the receptive field of convolution is inevitably limited, so these solutions are still incapable of effectively capturing sufficient global context to deal with the challenges mentioned above.
Recently, vision transformers have been proposed to regard an image as a sequence of patches and aggregate features in a global manner by self-attention mechanisms~\cite{carion2020detr,prangemeier2020attention,wang2021pyramid,wang2021pvtv2}. 
It is also verified that transformers can be used to handle medical image segmentation tasks, i.e., TransUNet~\cite{chen2021transunet} and TransFuse~\cite{zhang2021transfuse}.
In the field of skin lesion segmentation, studies improve the transformer-based networks with boundary information~\cite{wang2021boundary,cao2022icl}, while they have not thoroughly explored the potential usefulness of boundary information and global context in a multi-scale manner. 
Furthermore, these transformers still contain convolutional modules that may decrease the performance thanks to the inductive bias.

In this paper, we propose a novel cross-scale boundary-aware transformer (\textbf{XBound-Former}) to ably handle the problems mentioned above by holistically leveraging the advancement of boundary-wise prior knowledge and self-attention mechanism.
This method is inspired by the intuition that human beings perceive lesions in vision, i.e., considering global context to coarsely locate lesion areas and paying particular attention to the ambiguous area to specify the exact boundary.  
Concretely, we enhance the boundary modeling ability of the transformer-based network via three key learners: implicit boundary learner (im-Bound), explicit boundary learner (ex-Bound), and cross-scale boundary learner (X-Bound).
\begin{itemize}
    \item \textbf{Im-Bound} is recommended to explore local contexts for accurate boundary modeling implicitly. As the points with large boundary variation contribute more to the segmentation result than other boundary points, we constrain the network attention on such boundary key points.
    It enhances the local boundary modeling while maintaining the global context.
    
    \item \textbf{Ex-Bound} is proposed to explicitly extract the boundary knowledge as multiple embeddings where each embedding represents the boundary knowledge at a unique scale. They are used to further enhance the local boundary modeling and boost the cross-scale communication.
    
    \item \textbf{X-Bound} is suggested as a cross-scale attention mechanism for simultaneously addressing the problems of ambiguous boundaries and size variation. Acting like human beings that determine the accurate boundaries by zooming in and zooming out, we use the learned boundary embedding at one scale to guide the boundary-aware attention at the other scales to enhance the cross-scale knowledge communication.
\end{itemize}

We evaluate our model on two skin lesion datasets, ISIS-2016\&PH2 and ISIC-2018, following the standard experimental setup~\cite{9157193,wang2021boundary,cao2022icl}. 
To evaluate the generalization, we perform an extensive experiment on the polyp lesion which has closed characteristics.
Our model has achieved superior performance in all experiments compared to state-of-the-art CNN-based and transformer-based models, indicating the advanced power in addressing object segmentation with ambiguous boundaries, especially for skin lesion segmentation.

\section{Related Work}
\begin{figure*}[t]
\centering
    \includegraphics[width=\textwidth]{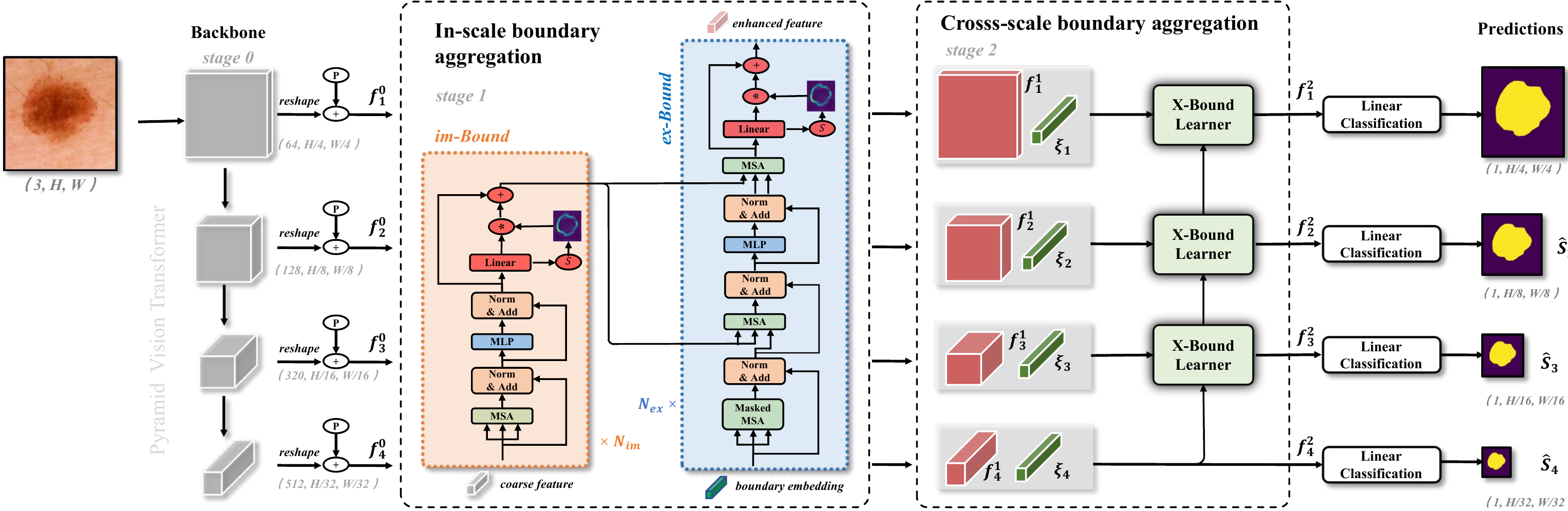}
    \caption{
    An overview of the proposed cross-scale boundary-aware transformer (XBound-Former). Briefly, it enhances the boundary modeling of transformers via in- and cross-scale boundary aggregation. Given coarsely extracted features ($\{f_l^0\}_{l=1}^{4}$), (1) the in-scale boundary aggregation implicitly (im-Bound) and explicitly (ex-Bound) explores boundary knowledge, enhancing features ($\{f_l^1\}_{l=1}^{4}$) and obtaining boundary embeddings ($\{\xi\}_{l=1}^{4}$) at each scale, and (2) the cross-scale boundary aggregation further exploits boundary information by fusing learned boundary embeddings from different scales and yielding final enhanced features ($\{f_l^2\}_{l=1}^{4}$). After that, several classification heads are used to predict the segmentation maps.
    }
    \label{fig:framework}
    \vspace{-4mm}
\end{figure*}
\subsection{Skin Lesion Segmentation}
In the early years, traditional methods apply various hand-crafted features to learn lesion segmentation that are not robust and stable.
It leads to poor segmentation performance when facing large lesions with large variations~\cite{5342426}.
Later, a fully convolutional network (FCN)~\cite{long2015fully} brings the deep learning model to skin lesion segmentation and achieves a much better result.
Several improved networks following its direction are proposed to solve the imbalance between foreground and background pixels~\cite{yuan2017automatic} and enhance the multi-scale feature representation~\cite{li2018dense}.
With the widespread use of the attention-based mechanism, channel and spatial attention-based methods are applied to enhance the lesion modeling~\cite{azad2020attention, gu2020net}.
The performance indeed reaches a higher score, but skin lesions' ambiguous boundaries are still hard to recognize.
To address this issue, ~\cite{wu2020automated} propose adaptive dual attention modules to let the network focus on lesion boundaries while it fails to cope with blurry boundaries owing to poor use of boundary-aware prior knowledge.
More recently, seeing the excellent success achieved by vision transformers, several studies employ transformer-based networks in the field of skin lesion segmentation~\cite{wang2021boundary,cao2022icl}.
It works to solve the problem of large lesion variation by capturing the global context.
However, they are still unable to handle the problem of ambiguous boundaries, especially the ones with size variation.
Instead, our proposed \textbf{XBound-Former} exploits multi-scale boundary information through the advanced self-attention blocks and utilizes the boundary-aware prior knowledge to supervise the transformer training. Thus it can outperform the State-of-the-Arts and the latest vision transformers.

\subsection{Vision Transformers}
Transformer, as a standard model in natural language process~\cite{vaswani2017attention}, has made great progress in the field of computer vision recently.
The first vision transformer, ViT~\cite{dosovitskiy2020vit}, proposes to split an image into a certain number of patches and utilize self-attention blocks to embed the features, achieving competitive performance in image classification tasks compared to the latest convolution-based neural networks.
Later work~\cite{touvron2021training} introduces a series of strategies to increase the training efficiency and improve the accuracy on small datasets.
Although the transformers are originally proposed to explore global dependency, recent studies find that the transformers also need local communication~\cite{liu2021swin,yuan2021tokens,wang2021pvtv2}, which can be achieved through the local window shift or pyramid architecture~\cite{wang2021pvtv2}, especially for the tasks requiring dense representations~\cite{carion2020end,zheng2021rethinking,liu2021swin}.
As for medical image segmentation, the effectiveness of vision transformers is verified by TransUNet~\cite{chen2021transunet} and TransFuse~\cite{zhang2021transfuse}.
In the field of skin lesion segmentation, vision transformers also boost the performance to reach new higher scores~\cite{wang2021boundary,cao2022icl}.
Despite their success, these models have not considered the complementary knowledge of boundary knowledge and global context in a multi-scale manner, which may help segment the extremely challenging lesions.
\textbf{XBound-Former} aims to mitigate this issue through cross-scale boundary learners and, besides, builds a pure attention-based network instead of the fusion of transformer and convolution to prevent the inductive bias.

\subsection{Boundary-aware Prior Knowledge}
The accurate recognition of ambiguous boundaries is one of the most tricky problems in medical image segmentation. There are plenty of works to address this issue by taking full advantage of the boundary-aware prior knowledge.
The earliest works propose to modify the loss function to give boundary-aware supervision for network optimization, i.e., HD loss~\cite{karimi2019reducing}, Boundary loss~\cite{kervadec2019boundary}, etc.
Later, multi-task learning is applied in this direction where manually designed tasks are used to provide extra supervision on the boundaries~\cite{wang2019ct, meng2021graph}.
Apart from the boundary-aware supervision, several networks propose to utilize spatial attention mechanisms to enhance the representation of boundaries ~\cite{wu2020automated}.
By contrast, we not only introduce the boundary-aware prior knowledge into vision transformers but also present a novel key-patch map generator that can select the most ambiguous points among the boundaries and convert them to the key-patch map to give supervision to the transformers.
\section{Method}
An overview of the cross-scale boundary-aware transformer (\textbf{XBound-Former}) is presented in Fig.~\ref{fig:framework}, where we show the details about how to leverage boundary prior knowledge and global dependency across different scales. 
It first utilizes a pyramid vision transformer~\cite{wang2021pyramid} to coarsely extract the features of an input dermoscopy skin image.
As a pyramid feature extractor, the backbone yield features at four different scales, $\{f_l^0\}_{l=1}^4$.
Here, $f_1^0$ denotes the lowest feature with the largest scale and $f_4^0$ denotes the deepest feature with the smallest scale ($f_{l}^0 \in \mathbb{R}^{C^{l}\times\frac{H}{2^{l+1}}\times\frac{W}{2^{l+1}}}$).
Each feature will be enhanced through the in-scale and cross-scale boundary aggregation to strengthen the boundary representation.
Finally, several linear classification heads are used to predict the segmentation maps.

\subsection{In-scale Boundary Modeling}
As an attention-based mechanism, transformers treat each image as a sequence of patches and explore the global dependency to represent them.
The global view is precisely helpful for the vision tasks, while recent studies have shown that they also require local context modeling in the dense-level vision tasks~\cite{liu2021swin,wang2021pyramid}.
For the segmentation tasks, especially for skin lesions with ambiguous boundaries, global dependency can help locate coarse boundary but lacks local contexts to segment accurate boundaries.
Therefore, we propose to fuse boundary information in the transformers to explore the local context of boundaries.
It is achieved by using a sequence of $N_{im}$ implicit boundary learners (im-Bound) and $N_{ex}$ explicit boundary learners (ex-Bound) to refine the feature at each scale as $\{f_l^1\}_{l=1}^4$,
where $f_{l}^1 \in \mathbb{R}^{C^{l}\times\frac{H}{2^{l+1}}\times\frac{W}{2^{l+1}}}$.
The process is denoted as,
\begin{equation}
    f^{1}, \xi \longleftarrow \mathcal{F}_{\textit{ex-Bound}}^{1...N_{ex}}\;(\mathcal{F}_{\textit{im-Bound}}^{1...N_{im}}\;(f^0)),
\end{equation}
where we simplify the notation $l$.
As the in-scale boundary modeling module takes the sequential features instead of 2-D maps as inputs and outputs, we re-define the inputted features as $z_l \in \mathbb{R}^{C^{l}\times\frac{H*W}{4^{l+1}}}$.
They are the encoded features after sequentialization and are added with position embeddings~\cite{gehring2017convolutional}.

\subsubsection{Implicitly Boundary-wise Attention}
The \textbf{im-Bound} aims to constrain the model's attention on the points with large boundary variation as they contribute more to the final segmentation result.
With this inspiration, we propose to utilize the self-attention module to find such points in the manner of predicting boundary key-point map.
The map is used for the feature refinement and offering a boundary-aware constraint.
Specifically, it contains $N_{im}$ cascaded blocks in total.
Assumed that at the $i$-th block, given the inputted feature as $z^{i-1}$, where $z^0\leftarrow z$, we firstly feed it into a sequence of multi-head self-attention (MSA) and multi-layer perception (MLP) to gather the global dependency for coarsely locating the boundaries~\cite{dosovitskiy2020vit}.
After each part, there is a Layer Normalization with residual short connection for a stable training process~\cite{vaswani2017attention}.
We denote this intermediate feature as,
\begin{equation}
\label{eq:msa}
    \rho^{i} = \mathcal{F}_{MSA}(z^{i-1}) \oplus \mathcal{F}_{MLP}(\mathcal{F}_{MSA}(z^{i-1})),
\end{equation}
where $\oplus$ denotes the element-wise addition and $\mathcal{F}_{MSA}(\textit{query},\textit{key},\textit{value})$ denotes the MSA operation.
As the self-attention modules embed \textit{query}, \textit{key} and \textit{value} together from $z^{i-1}$, we simplify the equation.
Additionally, the LayerNorm operation is also simplified to save space.
Then, a linear predictor with Sigmoid activation is utilized to classify each patch whether it is the point with large boundary variation, supervised by the boundary key-point map pre-produced by our boundary key-point map generation algorithm (see Sec.~\ref{sec:generation}).
We denote the predicted key-point map as $\hat{M}^{i}$ so that we could obtain the enhanced feature as,
\begin{equation}
    z^{i} = \rho^{i}\ \oplus\ (\ \rho^{i} \otimes \hat{M}^{i}\ ),
\end{equation}
where $\oplus$ denotes the element-wise multiplication.
After $N_{im}$ cascaded blocks, the resulted feature $z^{N_{im}}$ will be sent to ex-Bound for further refinement.

 \begin{figure}[t]
    \centering
    \includegraphics[width=\linewidth]{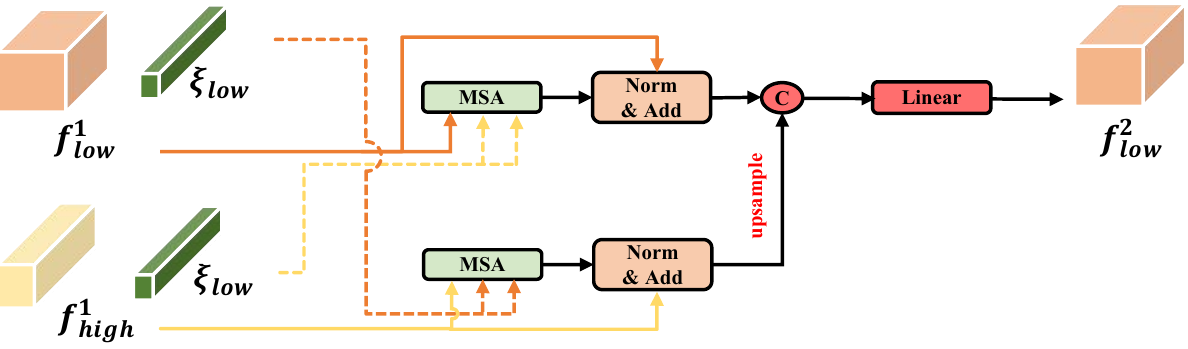}
    \caption{Details of the cross-scale boundary learner (x-Bound). It utilizes the boundary embedding at one scale to guide the multi-head self-attention (MSA) module at another scale. The enhanced features are combined and projected to keep the same dimension with the feature at a low level ($f_{low}^{1}$) via a linear projection.
    }
    \label{fig:xbound}
    \vspace{-3mm}
\end{figure}
 \begin{figure}[t]
    \centering
    \includegraphics[width=\linewidth]{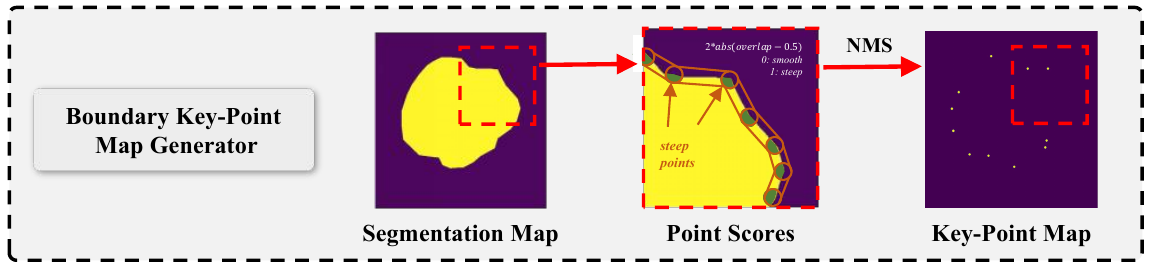}
    \caption{
    The pipeline of the boundary key-point map generation. It aims to transform the ground-truth segmentation map into the key-point map for supervising boundary learners.
    }
    \label{fig:generation}
    \vspace{-4mm}
\end{figure}
\subsubsection{Learn Explicit Boundary Embedding}
The \textbf{ex-Bound} is proposed to embed boundary information into a set of feature vectors explicitly $\{\xi_l\}_{l=1}^4$, where each embedding contains the high-level boundary semantics at a unique scale.
This learner is different from the \textbf{im-Bound} regarding the implementation, as well as the motivation that it not only refines the features but also provides the explicit expression for subsequent cross-scale communication.
To achieve this goal, we treat the boundary key points as query objects and employ a transformer decoder~\cite{carion2020end,prangemeier2020c} to learn the boundary embeddings.
The decoder contains a sequence of the Masked MSA module, MSA module, and the MLP module, each after which there is a LayerNorm layer and the short connection~\cite{carion2020end}.
Thanks to the global context modeling, it refines the inputted randomly initialized vector into the boundary embedding that contains abundant boundary knowledge.
After that, we send the feature and boundary embedding into the MSA module and the boundary key-point prediction part for the consideration of refining features and, of more importance, obtaining a preciser boundary embedding.

We repeat the ex-Bound $N_{ex}$ times to guarantee the adequate boundary learning.
For the $j$-th block, it takes feature $z^{N_{im}+j-1}$ and current embedding $\xi^{j-1}$ as input and output the aggregated feature $z^{N_{im}+j}$, the embedding $\xi_{j}$, and predicted key-point map $\hat{M}^{N_{im}+j}$. 
After $N_{ex}$ blocks, the resulted feature $z^{N_{im}+N_{ex}}$ is reshaped as $f^{1}$ and sent to the cross-scale boundary aggregation along with the learned boundary embedding.

\subsection{\vspace{-1mm}Attention-based Cross-scale Boundary Fusion}
\label{sec:xbound}
Automatic skin lesion segmentation suffers from the significant variance in lesion size and ambiguous boundaries. 
We take the first attempt to address these two issues simultaneously through the attention-based mechanism, our cross-scale boundary learners (\textbf{X-Bound}).
It is inspired by the human beings that determine the accurate boundaries by zooming in and zooming out boundaries and combining multi-perspective information across different scales to make the final decision.  

Generally, we visualize the details in Fig.~\ref{fig:xbound} where features and boundary embeddings at low scale ($f_{low}^1 \in \mathbb{R}^{C \times H_{low} \times W_{low}}, \xi_{low} \in \mathbb{R}^{1 \times C}$) and high scale ($f_{high}^1 \in \mathbb{R}^{C \times H_{high} \times W_{high}}, \xi_{high} \in \mathbb{R}^{1 \times C}$) are inputted and the enhanced feature at low scale ($f_{low}^2$) is outputted.
$(H_{low}, W_{low})$ denotes the size larger than $(H_{high}, W_{high})$.
Theoretically, the boundary embedding at a lower scale focuses on more local details and the boundary embedding at a larger scale focuses more on the high-level semantics.
Thus, utilizing the embedding at one scale to attentively refine the features at another scale provides complementary boundary knowledge.

In detail, we compare $\xi_{high}$ to each point in the lower feature $f_{low}^1$ and compute the distance matrix, which is then used to transfer boundary knowledge in $\xi_{high}$ to each point in the feature $f_{low}^1$.
It means that the intermediate features can be calculated as:
\begin{align}
    \gamma_{low} &= f_{low}^1 \oplus \mathcal{F}_{MSA}\ (f_{low}^1\ ,\ \xi_{high},\ \xi_{high}), \\ 
    \gamma_{high} &= f_{high}^1 \oplus \mathcal{F}_{MSA}\ (f_{high}^1\ ,\ \xi_{low}\ ,\ \xi_{low}),
\end{align}
where $\mathcal{F}_{MSA}$ is the multi-head attention module used in Equation~\ref{eq:msa}.
After that, the intermediate features are concatenated after the up-sample operation of $f_{high}^1$, which is fed into a linear projection head to reduce the feature dimension and refine the fusion. 
The resulted feature is denoted as $f_{low}^2$.

Totally, except the deepest feature $f_4^{1}$, we perform the cross-scale boundary learning on $\{f_{l}^{1}\}_{l=1}^3$ to obtain $\{f_{l}^{2}\}_{l=1}^3$ and $\{f_{4}^{2}\}$ is straightly set as $\{f_{4}^{1}\}$. 
For the consideration of multi-scale model learning, we feed each feature into a linear classification head to predict the segmentation maps $\{\hat{S}_{l}\}_{l=1}^4$.

\subsection{Boundary Key-point Generation Algorithm}
\label{sec:generation}
As the boundary learners do not naturally know which points can best represent the ambiguous boundaries, we propose a novel generation algorithm to pre-produce a ground-truth key-point map supervising the boundary learning, as shown in Fig.~\ref{fig:generation}.
The first step is to calculate all points on the boundary using a conventional contour detection algorithm~\cite{suzuki1985topological}.
After that we could obtain a set of coordinates of the boundary points.
Then, as points with larger boundary deviation should be paid more attention to than those with smoother deviation, we propose filtering the points by scoring the deviation.
For each point in this set, we draw a circle of radius $r$ and calculate the proportion $p$ of the lesion area in this circle region, where the larger or smaller $p$ indicates that the boundary is not smooth in this circle region.
Hence, we score each point as $|p-0.5|$ to representation its deviation.
To find the most valuable points, mon-maximum suppression is performed in which the points with larger $p$ than neighbor $k$ points are selected.
Next, selected points' 2D locations are mapped into the binary key-point map $M$, where points at the selected location are set to one and others are set to zero.
By minimizing the error between $M$ and $\hat{M}$, the supervision helps the boundary learners focus on the ambiguous boundary regions and helps the boundary embeddings learn correct boundary knowledge.

\subsection{\vspace{-1mm}Objective Function}
\label{section_objective_function}
We design a joint objective to train the entire network, including the lesion segmentation loss $L_{Seg}$ for predicted segmentation maps and the key-point map loss $L_{Map}$ for predicted boundary key-point maps, as
\begin{align}
\label{eq:loss}
    L_{Seg} &= \frac{1}{4}\sum_{l=1}^{4} \phi_{Dice}(\hat{S}_{l}, S_{l}), \\ 
    L_{Map} &= \frac{1}{N_{im}+N_{ex}}\sum_{l=1}^{4} \phi_{CE}(\hat{M}_{l}, M_{l}), \\
    L_{Total} &= L_{Seg} + \lambda L_{Map},
\end{align}
where $\phi_{DICE},\phi_{CE}$ denote Dice loss function and Cross Entropy function and $\{S_{l}\}_{l=1}^4,\{M_{l}\}_{l=1}^4$ are the ground-truth segmentation and boundary key-point maps pre-produced. 
$\lambda$ is the weight to balance the two objectives.
The detailed calculation is described as,
\begin{equation}
    \begin{aligned}
    \phi_{Dice}(\hat{S}, S) &= 1\ -\ 2*|S|*|\hat{S}|\ /\ (|S|+|\hat{S}|), \\
    \phi_{CE}(\hat{M}, M) &= -\hat{M}\log(M) - (1-\hat{M})\log(1-M).    
    \end{aligned}
\end{equation}

For deeply multi-scale supervision, given the original segmentation label, $S \in \mathbb{R}^{1 \times H \times W}$, we repeat the down-sample operation with different rates to obtain the set of ground-truth segmentation maps $\{S_{l}\}_{l=1}^4$, where $S_{l} \in \mathbb{R}^{1 \times \frac{H}{2^{l+1}} \times \frac{W}{2^{l+1}}}$. 
For the key-point maps, we also repeat the down-sample operation and obtain $\{M_{l}\}_{l=1}^4$ where $M_{l} \in \mathbb{R}^{1 \times \frac{H}{2^{l+1}} \times \frac{W}{2^{l+1}}}$. 

\section{Experiments}
\subsection{Dataset}
Following the classical experimental setting in the previous studies~\cite{9157193}, we evaluate our model on two skin lesions segmentation datasets, \textbf{ISIC-2016\&PH$^2$} and \textbf{ISIC-2018}. 
To further evaluate the model generalization, we evaluate it on the polyp lesion segmentation using five public polyp image datasets, named \textbf{Polyp-seg}.
\begin{itemize}
    \item The \textbf{ISIC-2016\&PH$^2$} contains samples from two centers to evaluate the accuracy and generalization ability of skin lesion segmentation. One is the ISIC-2016 dataset that contains a total number of 900 samples for training and 379 samples for validation. The other one is the PH$^{2}$ dataset~\cite{mendoncca2013ph}, containing 200 samples in total. Here, we use samples in the ISIC-2016 dataset for model learning through the official \textit{train}-\textit{validation} split and test the model on the 200 samples from the PH$^2$ dataset.
    
    \item The \textbf{ISIC-2018} dataset was also collected by ISIC in 2018, which contains 2594 images and labels. The resolution of each image varies from $720\times540$ to $6708\times4439$. As the public test set has not been released, we perform a 5-fold cross-validation for a fair comparison.
    
    \item The \textbf{Polyp-seg} dataset is collected following the most popular setting~\cite{fan2020pra}, which contains five public datasets: Kvasir-SEG~\cite{jha2020kvasir}, ClinicDB~\cite{bernal2015wm}, ColonDB~\cite{tajbakhsh2015automated}, Endoscene~\cite{vazquez2017benchmark}, and ETIS~\cite{silva2014toward}. 
    The Kvasir-SEG and ClinicDB contain  612 and 1000 samples, respectively, of which 548 and 900 samples are used for training and the rest samples are used for testing. 
    To evaluate the generalization ability, samples from the rest three datasets are also used for testing.
\end{itemize}

\begin{table*}[ht]
    \caption{Comparison of skin lesion segmentation with different approaches on the ISIC-2016\&PH$^{2}$ dataset. We report the averaged scores of the ISIC-2016 validation set and the scores of the PH$^{2}$ test set.}
\centering
\renewcommand\arraystretch{1.2}
\setlength\tabcolsep{10pt}
\begin{tabular}{c|cccc|cccc}
\hline
\multirow{2}{*}{Method} 
& \multicolumn{4}{c|}{\textit{validation}-ISIC-2016~\cite{gutman2016skin}}
& \multicolumn{4}{c}{\textit{test}-PH$^2$~\cite{mendoncca2013ph}} 
\\
\cline{2-9}
& IoU$\uparrow$ & Dice$\uparrow$ & ASSD$\downarrow$ & HD95$\downarrow$
& IoU$\uparrow$ & Dice$\uparrow$ & ASSD$\downarrow$ & HD95$\downarrow$\\
\hline
U-Net~\cite{ronneberger2015u}
& 80.25& 87.81& 15.51& 45.88
& 73.91& 83.66& 21.50& 60.12
\\
U-Net++~\cite{zhou2018unetpp}
& 81.84& 88.93& 15.01& 44.83
& 81.26& 88.99& 15.97& 46.66
\\
Attention U-Net~\cite{oktay2018attention} 
& 79.70& 87.43& 16.41& 48.78
& 69.52& 80.52& 26.73& 74.51
\\
DeepLabV3+~\cite{Chen_2018_ECCV}
& 85.62& 91.76& 9.85& 26.66
& 82.03& 89.56& 14.93& 37.81
\\
CE-Net~\cite{gu2019cenet}
& 84.39& 90.74& 11.77& 31.01
& 83.48& 90.44& 13.48& 33.97
\\
CA-Net~\cite{gu2020net} 
& 80.73& 88.10& 15.67& 44.98
& 75.18& 84.66& 21.06& 64.53
\\
\hline
TransFuse~\cite{zhang2021transfuse} 
& 86.19& 92.03& 10.04& 30.33
& 82.32& 89.75& 15.00& 39.98
\\
TransUNet~\cite{chen2021transunet}
& 84.89& 91.26& 10.63& 28.51
& 83.99& 90.96& 12.65& 33.30
\\
\hline
XBound-Former (Ours)
& \textbf{87.69}& \textbf{93.08}& \textbf{8.21}& \textbf{21.83}
& \textbf{85.38}& \textbf{91.80}& \textbf{10.72}& \textbf{26.00}
\\
\hline
\end{tabular}
\label{tab:results_2016}
\end{table*}

\begin{table*}[ht]
    \caption{Comparison of skin lesion segmentation with different approaches with 5-fold cross-validation on ISIC-2018 dataset. We present the averaged result and the standard error of all folds.}
    \label{exp}
\centering
\renewcommand\arraystretch{1.2}
\setlength\tabcolsep{5pt}
\begin{tabular}{c|cc|cc|cc|cc|cc|cc}
\hline
\multirow{2}{*}{Method} 
& \multicolumn{2}{c|}{Overall}
& \multicolumn{2}{c|}{Fold-1}
& \multicolumn{2}{c|}{Fold-2}
& \multicolumn{2}{c|}{Fold-3}
& \multicolumn{2}{c|}{Fold-4}
& \multicolumn{2}{c}{Fold-5}\\
\cline{2-13}
& IoU$\uparrow$ & ASSD$\downarrow$
& IoU$\uparrow$ & ASSD$\downarrow$
& IoU$\uparrow$ & ASSD$\downarrow$
& IoU$\uparrow$ & ASSD$\downarrow$
& IoU$\uparrow$ & ASSD$\downarrow$
& IoU$\uparrow$ & ASSD$\downarrow$\\
\hline
U-Net~\cite{ronneberger2015u}
& 75.40& 17.74
& 76.50& 16.66
& 77.84& 15.43
& 73.90& 19.76
& 74.38& 18.36
& 74.40& 18.50
\\
U-Net++~\cite{zhou2018unetpp}
& 77.92& 16.13
& 79.22& 13.83
& 79.28& 14.72
& 76.30& 19.24
& 77.07& 15.70
& 77.72& 17.14
\\
Attention U-Net~\cite{oktay2018attention} 
& 75.94& 17.24
& 78.01& 15.13
& 77.31& 16.04
& 73.78& 19.16
& 75.58& 17.15
& 75.01& 18.73
\\
DeepLabV3+~\cite{Chen_2018_ECCV}
& 82.49& 10.75
& 84.33& 8.57
& 84.05& 9.81
& 82.01& 10.90
& 82.57& 10.50
& 79.49& 14.00
\\
CE-Net~\cite{gu2019cenet}
& 82.86& 10.59
& 84.10& 8.83
& 83.75& 10.38
& 81.45& 12.12
& 82.81& 10.34
& 82.18& 11.26
\\
CA-Net~\cite{gu2020net} 
& 78.94& 13.52
& 80.56& 11.70
& 79.53& 13.21
& 77.88& 14.09
& 78.55& 14.06
& 78.20& 14.54
\\
\hline
TransFuse~\cite{zhang2021transfuse} 
& 83.59& 10.21
& 85.05& 8.80
& 84.45& 10.05
& 82.52& 11.07
& 83.29& 10.01
& 82.66& 11.13
\\
TransUNet~\cite{chen2021transunet}
& 82.61& 10.88
& 83.53& 9.72
& 83.90& 10.00
& 81.67& 11.86
& 82.37& 10.77
& 81.58& 12.06
\\
\hline
XBound-Former~(Ours)
& \bf84.51& \bf8.61
& \bf85.31& \bf7.67
& \bf85.22& \bf8.48
& \bf84.12& \bf8.92
& \bf84.00& \bf8.87
& \bf83.88& \bf9.13
\\
\hline
\end{tabular}
\label{tab:results_2018}
\end{table*}

\begin{table*}[ht]
    \caption{Comparison of polyp lesion segmentation. We report the overall $IoU$ and $F_{\beta}^{w}$ scores and the scores of each dataset.}
    \label{exp}
\centering
\renewcommand\arraystretch{1.2}
\setlength\tabcolsep{5pt}
\small
\begin{tabular}{c|cc|cc|cc|cc|cc|cc}
\hline
\multirow{2}{*}{Method} 
& \multicolumn{2}{c|}{Overall} 
& \multicolumn{2}{c|}{Kvasir-SEG} 
& \multicolumn{2}{c|}{ClinicDB} 
& \multicolumn{2}{c|}{ColonDB} 
& \multicolumn{2}{c|}{ETIS} 
& \multicolumn{2}{c}{Endoscene} \\
\cline{2-13}
& IoU$\uparrow$ & $F_{\beta}^{w}$ $\uparrow$
& IoU$\uparrow$ & $F_{\beta}^{w}$ $\uparrow$
& IoU$\uparrow$ & $F_{\beta}^{w}$ $\uparrow$
& IoU$\uparrow$ & $F_{\beta}^{w}$ $\uparrow$
& IoU$\uparrow$ & $F_{\beta}^{w}$ $\uparrow$
& IoU$\uparrow$ & $F_{\beta}^{w}$ $\uparrow$\\
\hline
ACSNet~\cite{zhang2020adaptive}&  65.07&  69.08&  83.80&  88.20&  82.60&  87.30&  63.10&  68.40&  49.60&  50.60&  78.80&  83.00\\
PraNet~\cite{fan2020pra}&  67.52&  72.41&  84.00&  88.50&  84.90&  89.60&  64.00&  69.90&  56.70&  60.00&  79.70&  84.30\\
MSEG~\cite{huang2021hardnet}&  66.68&  70.75&  83.90&  88.50&  86.40&  90.70&  64.90&  69.70&  50.90&  53.00&  80.40&  85.20\\
DCRNet~\cite{yin2022duplex}&  70.00&  74.95&  82.50&  86.80&  84.40&  89.00&  66.60&  72.40&  63.00&  67.10&  78.70&  82.50\\
EU-Net~\cite{patel2021enhanced}&  70.41&  74.55&  85.40&  89.30&  84.60&  89.10&  68.10&  73.00&  60.90&  63.60&  76.50&  80.50\\
SANet~\cite{wei2021shallow}&  71.38&  76.09&  84.70&  89.20&  85.90&  90.90&  67.00&  72.60&  65.40&  68.50&  81.50&  85.90\\
Polyp-PVT~\cite{dong2021PolypPVT}&  76.00&  81.60&  86.40&  \bf{91.10}&  88.90&  93.60&  72.70&  79.50&  70.60&  75.00&  83.30&  88.40\\
\hline
XBound-Former~(Ours)
&  \bf{77.50}&  \bf{84.30}&  \bf{87.10}&  89.70&  \bf{91.10}&  \bf{94.40}&  \bf{73.20}&  \bf{81.40}&  \bf{75.10}&  \bf{82.00}&  \bf{83.60}&  \bf{90.10}\\
\hline
\end{tabular}
\label{tab:results_polyp}
\end{table*}

\subsection{\vspace{-1mm}Evaluation Metrics}
We employ four widely-used metrics to quantitatively evaluate the skin lesion segmentation performances, including $Dice$ coefficient, $IoU$ score, Average symmetric surface distance ($ASSD$), and Hausdorff distance of boundaries (95$^{th}$ percentile; $HD95$).
Generally, a better segmentation performance shall have higher area-based metrics ($Dice, IoU$) and lower boundary-based metrics ($ASSD, HD95$).

The area-based similarity of predicted segmentation map $\hat{S}$ and the ground-truth $S$ are computed as:
\begin{equation}
    \begin{aligned}
        \psi_{Dice}(\hat{S}, S) &= 2*\frac{|\hat{S}*S|}{|\hat{S}|+|S|}, \\
        \psi_{IoU}(\hat{S}, S)&=\frac{|\hat{S}*S|}{|\hat{S}|+|S|-|\hat{S}*S|}.
    \end{aligned}
\end{equation}

To better evaluate the segmentation performance of boundaries, we employ another two boundary-based metrics, as
\begin{equation}
    \begin{aligned}
    \psi_{ASSD}(\hat{S}, S)&= \frac{\sum_{a \in P_{b}} d\left(a, G_{b}\right)+\sum_{b \in G_{b}} d\left(b, P_{b}\right)}{\left|P_{b}\right|+\left|G_{b}\right|},\\
    \psi_{HD95}(\hat{S}, S)&= \max \left\{h(P_{b}, G_{b}),h(G_{b}, P_{b})\right\},
    \end{aligned}
\end{equation}
where $P_{b}$ and $G_{b}$ denote the predicted boundary points and the ground-truth boundary points in the $\hat{S}$ and $S$, and $d(\cdot)$ denotes the minimum Euclidean distance function.
Moreover, $h(P_{b}, G_{b})=\hat{\underset{a \in P_{b}}{\max}} \left\{\underset{b \in G_{b}}{\min} \left\|a-b\right\|\right\}$ denotes the one-way hausdorff distance from $P_{b}$ to $G_{b}$, and $\hat{\max}\left\{\cdot\right\}$ refers to the calculation of the $95^{th}$ percentile of the distances.

As for the polyp segmentation, we adopt the same metrics as the latest work, Polyp-PVT~\cite{dong2021PolypPVT}, including the area-based metric, IoU, and the boundary-based metric, $F_{\beta}^{w}$.

\subsection{Implementation Details}
All methods are implemented on the Pytorch with a single NVIDIA Geforce GTX 3090 GPU with a memory of 24 GB. 
We empirically resize all images to $(512\times512)$ considering the computation efficiency.
A series of data augmentations are implemented to increase the data diversity, including vertical flip, horizontal flip, and random scale change (limited 0.9-1.1).
Each mini-batch includes eight images, and the AdamW~\cite{loshchilov2018fixing} optimizer with an initial learning rate of 0.0003 is used to optimize the parameters. 
We train the network for 200 epochs and save the model parameters with the best performance during validation.
We adopt the pyramid vision transformer, PVTv2~\cite{wang2021pvtv2}, as the backbone and pre-train it on the ImageNet dataset.
As for the hype-parameters, we set $N_{im}$ and $N_{ex}$ to 2 by default and discuss it in Section~\ref{sec:n_im_ex}.
In the boundary key-point generation algorithm, considering the image size and lesion size, we set $r=2$ and $k=30$ by default.

 \begin{figure*}[t]
    \centering
    \includegraphics[width=.8\textwidth]{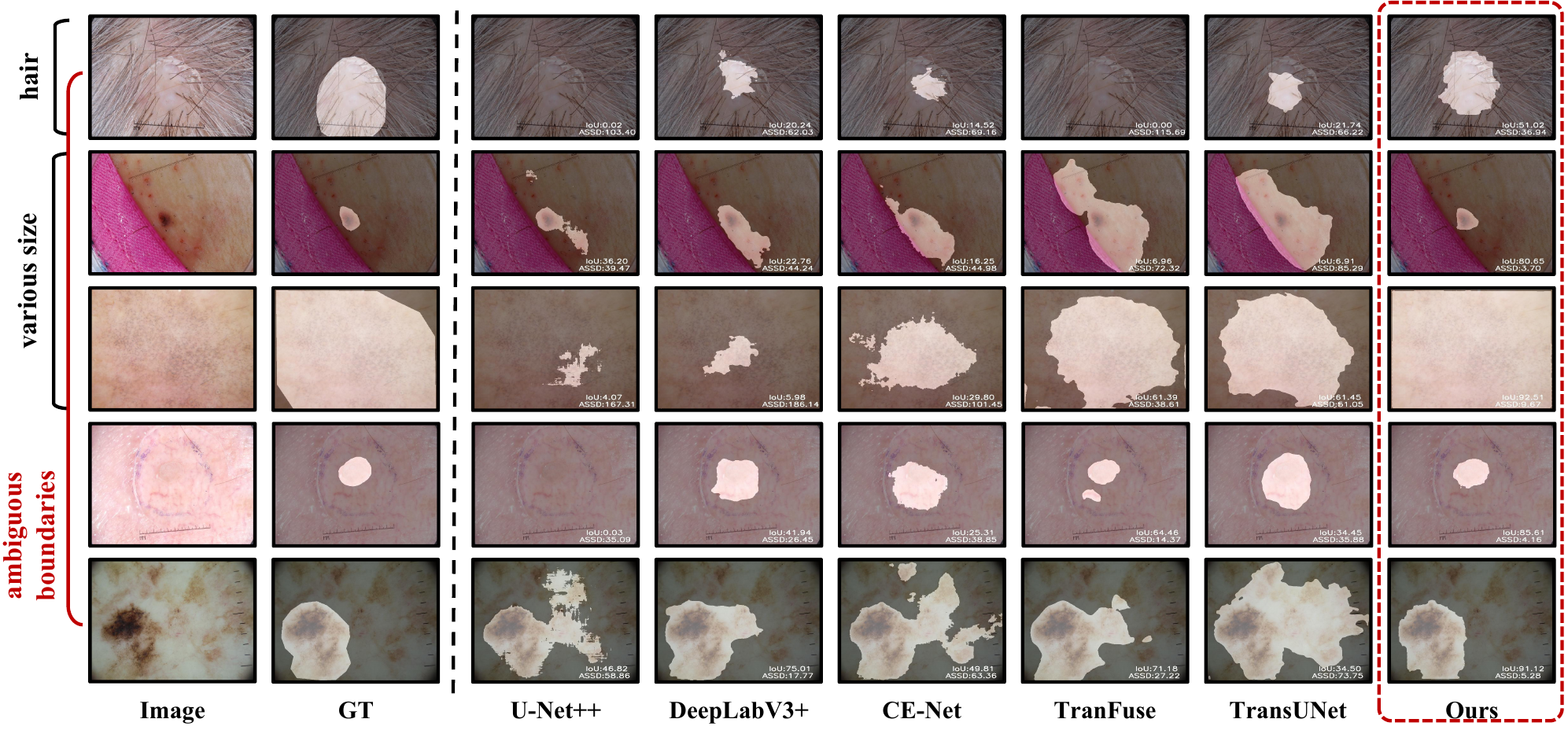}
    \vspace{-2mm}
    \caption{Visual comparison on some representative and challenging images. It includes the tricky lesions caused by hair occlusion, size variance, and especially the ambiguous boundaries. We show the specific IoU and ASSD scores in each visualized imag
    e at the right-bottom corner.
    }
    \label{fig:visual}
    \vspace{-2mm}
\end{figure*}

 \begin{figure*}[t]
    \centering
    \includegraphics[width=.9\textwidth]{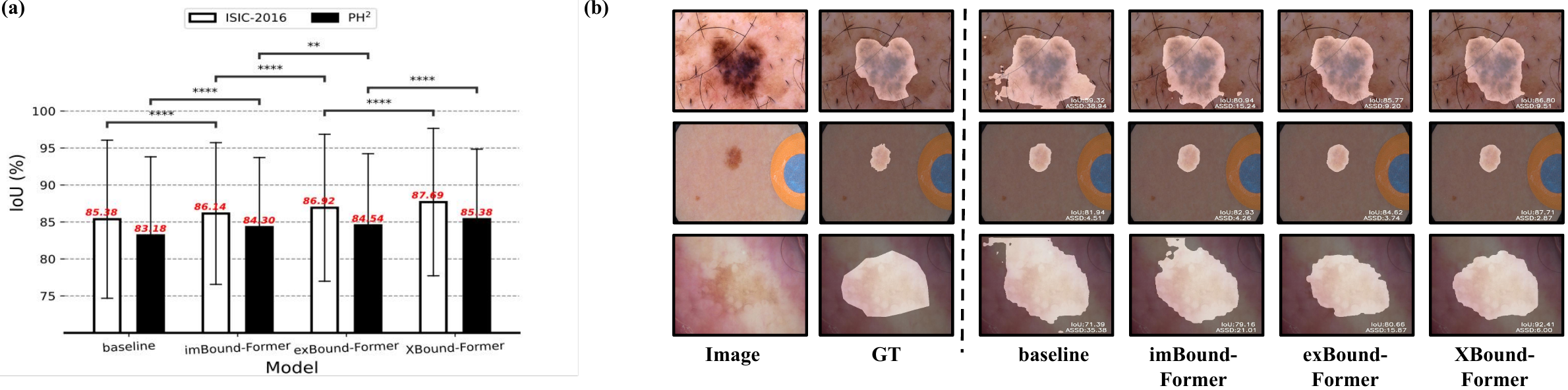}
    \vspace{-2mm}
    \caption{Ablation analysis of each boundary learner. 
    (a) Quantitative analysis of the ISIC-2016 validation set and PH$^2$ test set. The specific IoU score on each set is shown upper the plotted bar. We perform Wilcoxon test to analyze the distribution.
    (b) Visual comparison of some representative images. We show the specific IoU and ASSD scores at the right-bottom corner of the visualized image.
    }
    \label{fig:visual_ablation}
    \vspace{-4mm}
\end{figure*}

\subsection{Comparisons with state-of-the-art Methods}
\subsubsection{Quantitative results for skin lesion segmentation}
We compare our model to several popular segmentation models, including the CNN-based models, U-Net~\cite{ronneberger2015u}, U-Net++~\cite{zhou2018unetpp}, Attention U-Net~\cite{oktay2018attention}, DeepLabV3+~\cite{Chen_2018_ECCV}, CE-Net~\cite{gu2019cenet}, CA-Net~\cite{gu2020net}, and the transformer-based models, TransFuse~\cite{zhang2021transfuse} and TransUNet~\cite{chen2021transunet}.
All models are trained under the same setting as our model.

For the ISIC-2016\&PH$^2$ dataset, it is found that our model has achieved the best performance on whatever the validation set or the test set. Since the samples from the PH$^2$ dataset are unseen during the model learning, our superior performance indicates the satisfactory generalization ability, which is owing to the learning of boundaries that are the general features among different distributions. 
In comparison to us, TransFuse generalize poorly to the test set and TranUNet has poor segmentation accuracy on the validation set.
Furthermore, it is seen that our model has obviously lower $ASSD$ ($-1.83$ and $-1.93$) and $HD95$ ($-6.68$ and $-7.30$), demonstrating the promising advantage in handling boundary segmentation.

To extensively evaluate the models, we perform the 5-fold cross-validation in the ISIC-2018 dataset and show the evaluated scores of each fold as well as the overall scores in Table~\ref{tab:results_2018}.
The results illustrate that our model achieves the highest IoU score and the shortest ASSD distance on all sets.
In addition to this, although the improvement on the IoU score is not as large as that on the ISIC-2016\&PH$^2$ dataset, the ASSD score has decreased a lot compared to the other models.
It means that our model has superior accuracy in reducing the false positives away from the boundaries and detecting the ambiguous boundaries that are ignored by other models.

\subsubsection{Visualized Comparison for Skin Lesion Segmentation}
We visualize the predictions of some representative images in Fig.~\ref{fig:visual}, including the lesions with hair occlusion, various sizes, and ambiguous boundaries.
The first row shows that our model can detect the lesion covered by the hair with the largest accuracy.
The second and third rows prove that our model consistently yields stable and the best prediction on the smallest or largest lesions.
For all rows, particularly the last two rows where lesions show an extremely close appearance to neighbor tissues, our model is still able to give accurate segmentation.

 \begin{figure*}[t]
    \centering
    \includegraphics[width=.8\textwidth]{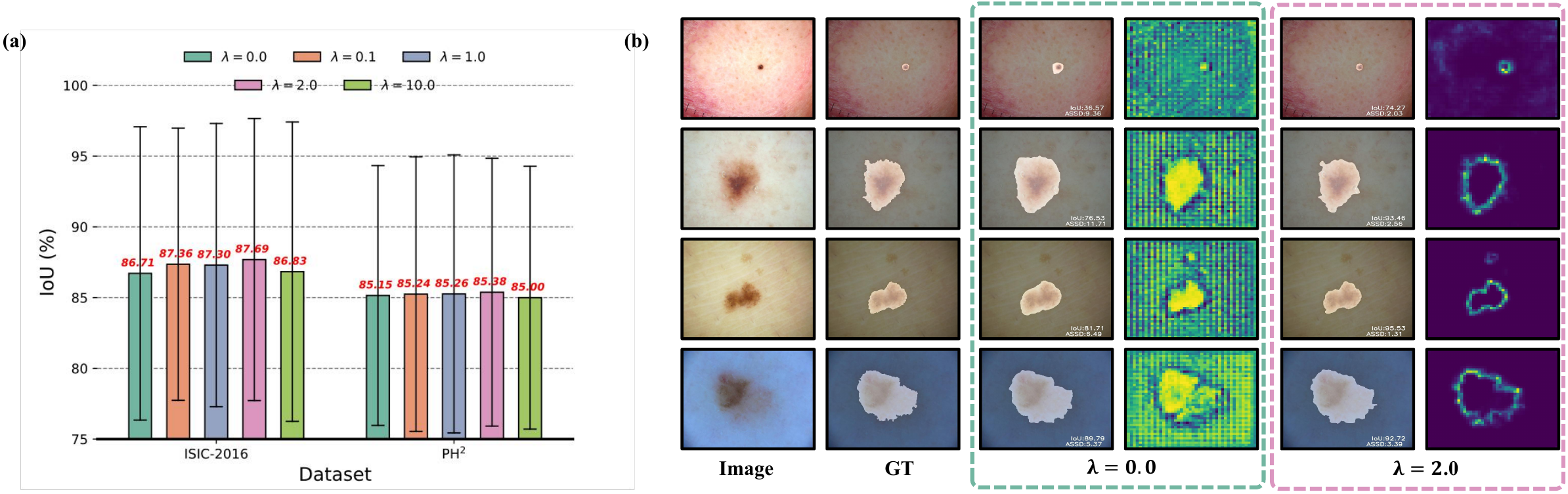}
    \caption{Analysis of how the boundary key-point supervises segmentation learning. (a) Evaluated scores with different controlling weight, $\lambda$. (b) Visual comparison of using key-point supervision or not.
    }
    \label{fig:bd_super}
    \vspace{-4mm}
\end{figure*}

\subsubsection{Extensive Evaluation for Polyp Segmentation}
We compare our model to the most popular polyp image segmentation models, including ACSNet~\cite{zhang2020adaptive}, PraNet~\cite{fan2020pra}, MSEG~\cite{huang2021hardnet}, DCRNet~\cite{yin2022duplex}, EUNet~\cite{patel2021enhanced}, SANet~\cite{wei2021shallow} and Polyp-PVT~\cite{dong2021PolypPVT}. 

We show the compared results in Table~\ref{tab:results_polyp}, where the overall scores and the scores of each dataset are presented.
We highlight the best score in bold, and it is found that our model nearly achieves the best scores on all metrics.
For overall performance, compared to the latest model, Polyp-PVT, which has also used PVTv2 as the backbone, our model yields obvious performance improvement, i.e., 1.5\% on the IoU score and 2.7\% on the $F_{\beta}^{w}$ score.
As $F_{\beta}^{w}$ demonstrates the ability of accurate boundary segmentation, the result indicates that our boundary learners are genuinely able to enhance the determination of boundary points.
The results on each dataset also support the conclusion, especially for the ETIS dataset.
Samples from the ETIS dataset are more challenging to segment, leading to relatively poorer performance in all experiments.
On such a difficult sampler, our model has a 4.5\% improvement on the IoU score and 7.0\% improvement on the $F_{\beta}^{w}$ score, indicating its superior ability to handle challenging boundaries.

\subsection{Analytical Ablation Study}
We conduct extensive ablation experiments on the ISIC-2016\&PH$^2$ dataset to demonstrate the effectiveness of the three bound learners in our proposed method. 
For the baseline comparison, we remove the learners of XBound-Former and maintain the same linear prediction and up-sampling fusion as U-Net.
Then, we add the im-Bound learners, ex-Bound learners, and X-Bound learners step by step and obtain three models that are the imBound-Former, exBound-Former, and XBound-Former.

\subsubsection{Quantitative Analysis}
The results of the ablation experiment are shown in Fig.~\ref{fig:visual_ablation}(a) using bar plots, and the evaluated IoU scores are highlighted by red scores.
Compared to the baseline model, imBound-Former has gained a $0.76\%$ improvement on the validation set and $1.12\%$ improvement on the test set, verifying its benefits in boosting segmentation accuracy and generalization.
In addition, exBound-Former gains further improvement on the validation set ($0.78\%$) and a slight improvement on the test set ($0.24\%$).
Since this module majorly aims to learn explicit embeddings for boundary knowledge that are essential for the X-Bound learners, the improvement is limited yet not important.
The complete version, XBound-Former, shows obvious and consistent improvement on the validation and test sets, verifying the usefulness of our attention-based cross-scale boundary fusion.

\subsubsection{Visual Comparison on Lesion Boundaries}
We also visually analyze the effectiveness of each component in Fig.~\ref{fig:visual_ablation}.
As it shows, the baseline model lacks sufficient ability to address lesions with ambiguous boundaries as there are a lot of false positives. 
This issue has decreased significantly in the predictions of imBound-Former and exBound-Former, while the determination is still not accurate enough.
By combining the multi-scale boundary knowledge, XBound-Former achieves the best performance on the small lesion (the second row) or the large lesion (the third row).

 \begin{figure}[t]
    \centering
    \includegraphics[width=\linewidth]{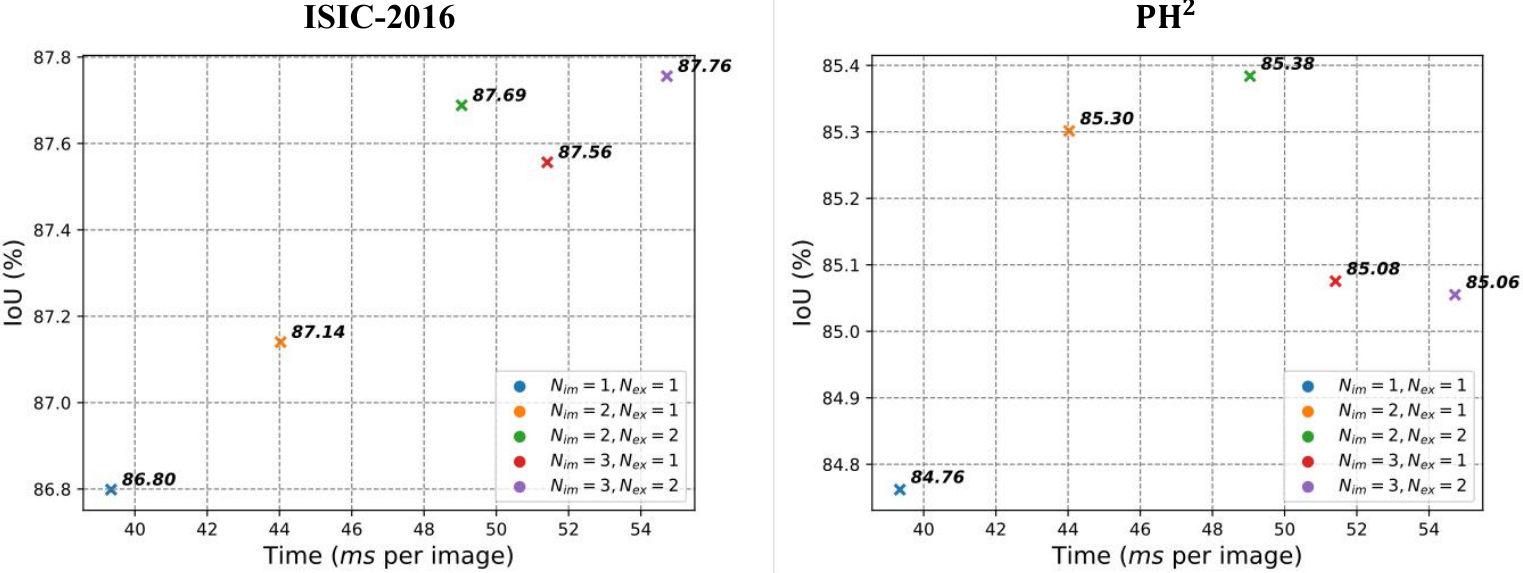}
    \caption{Inference speed analysis for models with different $N_{im}, N_{ex}$ on the ISIC-2016 validation set (left) and PH$^2$ test set (right). We measure the speed by calculating the inference time per image.
    }
    \label{fig:comp}
\end{figure}

\subsection{Detailed Analysis of Bound Learners}
\subsubsection{Boundary Supervision}
As shown in Equation~\ref{eq:loss}, we utilize the factor ($\lambda$) to balance the segmentation map loss and boundary key-point map loss. 
The smaller $\lambda$ may fail to provide strong enough supervision, while the larger $\lambda$ may sometimes bring the noise to the model learning.
Hence, we have a discussion about how it affects the final segmentation performance.
The results are shown in Fig.~\ref{fig:bd_super}, where $\lambda$ is set to $\{0, 0.1, 1, 2, 10\}$ and all models adopt the same architecture as XBound-Former.
As the plot shows in Fig.~\ref{fig:bd_super}(a), the evaluated scores increase on both sets when enlarging the $\lambda$ from $0.0$ to $2.0$.
However, they decrease when the $\lambda$ reaches $10.0$. 
It verifies the assumption that the small $\lambda$ limits the improvement and the large one will harm the segmentation training.
We additionally visualize the predicted segmentation map along with the point map in Fig.~\ref{fig:bd_super}(b).
As it shows, the model without boundary supervision is still able to predict coarse lesion regions for spatial attention while it lacks the ability to recognize the most challenging regions of the boundaries.
In comparison, our predicted point map concentrates on the ambiguous boundaries so that it can boost the challenging lesion's segmentation.

\subsubsection{Statistics of the Efficiency}
\label{sec:n_im_ex}
We set $N_{im}, N_{ex}$ to control the number of im-Bound and ex-Bound learners.
Enlarging them leads to more computation, while few learners may not be able to learn the correct boundary knowledge.
Fig.~\ref{fig:comp} shows the evaluated IoU scores and inference time of the models with different $N_{im}, N_{ex}$.
For the validation set, the evaluated IoU score increase obviously with more boundary learners, and the score changes a few when $N_{im} = 2, N_{ex}=2$. 
The IoU score also increases with increasing $N_{im} = 1, N_{ex}=1$ to $N_{im} = 2, N_{ex}=2$ but it also drops with $N_{im} = 3$.
The underlying reason may be that more learners bring larger hardness to model optimization.
Considering both the efficiency, accuracy and generalization ability, we take $N_{im} = 2, N_{ex}=2$ as our final setting.
\section{Discussion}
Skin lesion segmentation plays a vital role in the quantitative analysis of skin cancers, i.e., lesion size and shape analysis.
Existing studies adopt attention-based networks to catch global context, and boundary-aware supervision is proved to be effective for object segmentation in other fields.
In this work, we exploit the complementary advantage of global context and boundary knowledge at multi-scale, proposing a cross-scale boundary-aware transformer, XBound-Former, for precise segmentation of skin lesions with ambiguous boundaries.
The main contribution is our three boundary learners to explore in-scale and cross-scale boundary knowledge.
The experiment is conducted on two skin lesion datasets and an external polyp lesion dataset.
The results have shown that our model has the best segmentation performance, especially in the determination of challenging boundaries.
The generalization ability on unseen images and different tasks has also been verified.

In the medical field, targets usually have ambiguous boundaries that are hard to determine, even for human beings.
The challenges majorly come from the limitation of imaging techniques and would be solved in the future by the new evolution of advanced imaging techniques.
However, in the current community, how to segment these challenging objects has huge significance for the diagnosis, quality control, and treatment planning of patients.
Therefore, we thoroughly investigate and aim to solve the challenges in the skin lesion segmentation and preliminarily discuss the potential users on the other targets with similar characteristics.

How to fuse boundary information into the segmentation tasks is one of the most well-known topics in object segmentation. 
It can be achieved through designing boundary-aware loss objectives like HD loss. 
Recent studies show that it is more effective to transfer the boundary loss as boundary key-point map loss.
In addition to the supervision, the predicted boundary key-point map can also be used as the spatial attention map.
Following this direction, we propose XBound-Former, which takes the complementary usage of the attention-based network and boundary supervision.
Based on this theory, we further explore the potential help in exploring cross-scale boundary knowledge.
All our proposals are proved to be effective in our ablation experiment and the detailed discussion.

Our model still has some limitations that will further improve the segmentation if broken. First, in some extremely challenging images, the boundary key points are still unable to detect clearly.
The false point detection may bring harmful guidance to the branch of lesion segmentation. Although they have the complementary advantage in most cases, we should consider the potential harm in some noisy cases. Second, boundary key-point detection is a different task that requires unique representations compared to lesion segmentation. In future work, utilizing different models for the two branches instead of sharing the same architecture may be helpful to guarantee the accuracy of the two branches.
\section{Conclusion}
We present a novel cross-scale boundary-aware transformer for skin lesion segmentation, and it can be extended to similar targets that have ambiguous boundaries.
We perform comparison experiments on two skin lesion datasets to verify the segmentation accuracy and the generalization ability.
The extensive experiment conducted on the polyp segmentation also indicates our feasibility on more tasks.
The detailed ablation study proves that the improvement comes from our implicitly, explicitly, and cross-scale boundary modeling.
Besides, it is also found that our model still fails on some extremely low-contrast lesions, which may be solved by fusing a deep learning-based model and low-level feature extractor.

\section{Acknowledgement} 
This work is supported by the Ministry of Science and Technology of the People's Republic of China under grant No. 2021ZD0201900 and 2021ZD0201904.

\bibliographystyle{IEEETran}
\bibliography{refs}

\end{document}